# Un Algorithme génétique pour le problème de ramassage et de livraison avec fenêtres de temps à plusieurs véhicules

**I. Harbaoui Dridi**[(1),(2)]  **R. Kammarti**[(1),(2)]  **P. Borne**[(1)]  **M. Ksouri**[(2)]

imenharbaoui@gmail.com   kammarti.ryan@ec-lille.fr   p.borne@ec-lille.fr   Mekki.Ksouri@insat.rnu.tn

[(1)]LAGIS
Ecole Centrale de Lille
Villeneuve d'Ascq, France

[(2)] LACS
Ecole Nationale des Ingénieurs de Tunis
Tunis - Belvédère. TUNISIE

**Résumé :** Le problème de ramassage et de livraison (pick-up and delivery problem) est un problème d'optimisation de tournées de véhicules devant satisfaire des demandes de transport des biens entre fournisseurs et clients.

Il s'agit de déterminer un circuit de plusieurs véhicules, de façon à servir à coût minimal un ensemble de clients répartis dans un réseau, satisfaisant certaines contraintes relatives aux véhicules, à leurs capacités et à des précédences entre les nœuds. Dans cet article, nous présentons un état de l'art du PDPTW et nous proposons par la suite une approche basée sur les algorithmes génétiques permettant de minimiser la distance totale parcourue et par la suite le coût de transport, et ce en montrant un codage qui représente les paramètres caractérisant chaque individu.

**Mot Clés** : Tournées de véhicules, circuit de ramassage et de livraison, fenêtre de temps, optimisation, algorithme génétique.

## I. Introduction

La planification des tournées de véhicules constitue un enjeu logistique important tant aux niveaux des approvisionnements que du transport inter-usines ou le transport de distribution.

Dans le problème m-PDPTW auquel nous nous intéressons, nous considérons une flotte de véhicules $V_k$ de capacité $Q_k$ et un ensemble de biens à transporter des fournisseurs vers différentes destinations. L'objectif est de servir un ensemble de clients, sous certaines contraintes relatives aux véhicules, à leurs capacités et à des précédences entre les nœuds, et ce en minimisant le coût total de transport.

Dans cet article, nous présentons un état de l'art du PDPTW, suivie de l'approche proposée pour l'optimisation des circuits de ramassage et de livraison avec fenêtre de temps.

## II. Etat de l'art

Le problème général de construction de tournées de véhicules est connu sous le nom de *Vehicle Routing Problem* (VRP) et représente un problème d'optimisation combinatoire multi-objectif qui a fait l'objet de nombreux travaux et de nombreuses variantes dans la littérature. Il appartient à la catégorie NP-difficile. [1] [Christofides, N et al] [2] [Lenstra, J et al]

Des métaheuristiques ont été aussi appliquées pour la résolution des problèmes de tournée de véhicules. Parmi ces méthodes nous pouvons citer les algorithmes de colonies de fourmis, qui ont été utilisés par [3] [Montamenni, R et al] pour la résolution du DVRP.

Le principe du VRP (Figure 1) est, étant donné un dépôt D et un ensemble de commandes de clients C = {c1, . . ., cn}, de construire un ensemble de tournées, pour un nombre fini de véhicules, commençant et finissant à un dépôt. Dans ces tournées, un client doit être desservi une seule fois par un seul véhicule et la capacité de transport d'un véhicule pour une tournée ne doit pas être dépassée. [4] [Nabaa, M et al]

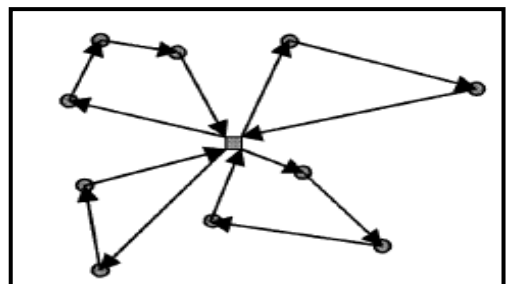

**Figure 1: Problème de tournées de véhicules**

Le problème de VRP avec des collectes et livraison de marchandises à fenêtres de temps correspond au PDPTW (*Pickup & Delivery Problem with Time Windows* ou Problème de Ramassage & Livraison avec Fenêtres de temps).

### A. Le PDPTW: Pickup and Delivery Problem with Time Windows

Le PDPTW est une variante du VRPTW où en plus de l'existence des contraintes temporelles, ce problème implique un ensemble de clients et un ensemble de fournisseurs géographiquement localisés. Chaque tournée doit également satisfaire les contraintes de précédence pour garantir qu'un client ne doit pas être visité avant son fournisseur. [5] [Psaraftis, H.N]

Cette variante de problème se divise en deux catégories : le 1-PDPTW (avec un seul véhicule) et le m-PDPTW (à plusieurs véhicules).

Une approche de programmation dynamique pour résoudre le 1-PDP sans et avec fenêtres de temps a été développée par [6] [H.N. Psaraftis] en considérant comme fonction objectif la minimisation d'une pondération de la durée totale de la tournée et de la non-satisfaction des clients.

[7] [Jih, W et al] ont développé une approche basée sur les algorithmes génétiques hybrides pour résoudre le 1-PDPTW, ayant pour objectif de minimiser une combinaison du coût total et de la somme des temps d'attente. L'algorithme développé utilise quatre différents opérateurs de croisement ainsi que trois opérateurs de mutation.

Dans d'autres travaux, un algorithme génétique a été développé par [8] [Velasco, N et al] pour résoudre le 1-PDP bi-objectif dans lequel la durée totale des tournées doit être minimisée tout en satisfaisant en priorité les demandes les plus urgentes. Dans cette littérature, la méthode proposée, pour résoudre ce problème est inspirée d'un algorithme appelé Non dominated Sorting Algorithm (NSGA-II).

D'autres travaux traitent le 1-PDPTW, en minimisant le compromis entre la distance totale parcourue, le temps total d'attente et le retard total et ce en utilisant un algorithme évolutionniste avec des opérateurs génétiques spéciaux, la recherche taboue et la Pareto optimalité pour fournir un ensemble de solutions viables. [9] [Kammarti, R et al] Ces travaux ont été étendus, et ce en proposant une nouvelle approche basée sur l'utilisation de bornes inférieures pour l'évaluation des solutions et de leur qualité, minimisant le compromis entre la distance totale parcourue et la somme des retards. [10]

En ce qui concerne le m-PDPTW, [11] [Sol, M et al] ont proposé un algorithme de *branch and price* pour résoudre le m-PDPTW et ce en minimisant le nombre de véhicules nécessaires pour satisfaire toutes les demandes de transport et la distance totale parcourue.

[12] [Quan, L et al] ont présenté une heuristique de construction basée sur le principe d'insertion ayant pour fonction objectif, la minimisation du coût total, incluant les coûts fixes des véhicules et les frais de déplacement qui sont proportionnels à la distance de déplacement.

Une nouvelle métaheuristique se basant sur un algorithme tabou intégré dans un recuit simulé, a été développé par [13] [Li, H et al] pour résoudre le m-PDPTW, ceci en recherchant, à partir de la meilleure solution courante s'il y a amélioration pendant plusieurs itérations.

[14] [Li, H et al] ont développé une méthode appelée « *Squeaky wheel* » pour résoudre le m-PDPTW avec une recherche locale.

### III. Modélisation Mathématique du m-PDPTW

Notre objectif est de servir toutes les demandes des clients, en minimisant le coût total de transport. Ce coût est relatif au nombre de véhicules utilisés et à la distance parcourue par chaque véhicule.

Notre problème est caractérisé par les paramètres suivants :

- N : Ensemble des nœuds clients, fournisseurs et dépôt,
- N' : Ensemble des nœuds clients, fournisseurs,
- N+ : Ensemble des nœuds fournisseurs,
- N− : Ensemble des nœuds clients,
- K : Nombre de véhicules,
- $d_{ij}$ : distance euclidienne entre le nœud i et le nœud j, si $d_{ij} = \infty$ alors le chemin entre i et j n'existe pas (impasse, rue piétonne,…)
- $t_{ijk}$ : temps mis par le véhicule k pour aller du nœud i au nœud j,
- $[e_i, l_i]$ : fenêtre de temps du nœud i,
- $s_i$ : temps d'arrêt au nœud i,
- $q_i$ : quantité à traiter au nœud i. Si $q_i > 0$, le nœud est fournisseur ; Si $q_i < 0$, le nœud est un client et si $q_i = 0$ alors le nœud a été servi.
- $Q_k$ : capacité du véhicule k,
- i = 0..N : indice des nœuds prédécesseurs,
- j = 0..N : indice des nœuds successeurs,
- k = 1..K : indice des véhicules,
- $Xijk = \begin{cases} 1 \textit{ Si le véhicule k voyage du noeud i vers le noeud j} \\ 0 \textit{ Sinon} \end{cases}$
- $A_i$ : temps d'arrivée au nœud i,

- $D_i$ : temps de départ du nœud i,
- $y_{ik}$ : quantité présente dans le véhicule k visitant le nœud i,
- $C_k$ : coût de transport associé au véhicule k,
- Un nœud (fournisseur ou client) n'est servi que par un seul véhicule et en une seule fois,
- Il y a un seul dépôt,
- Les contraintes de capacité doivent être respectées,
- Les contraintes de temps sont rigides concernant l'heure d'arrivée,
- Chaque véhicule commence le trajet du dépôt et y retourne à la fin,
- Un véhicule reste à l'arrêt à un nœud le temps nécessaire pour le traitement de la demande.
- Si un véhicule arrive au nœud i avant la date ei de début de sa fenêtre, il attend.

La fonction à minimiser est donnée comme suit :

$$Minimiser\ f = \left( \sum_{k \in K} \sum_{i \in N} \sum_{j \in N} C_k d_{ij} X_{ijk} \right) \quad (1)$$

Sous :

$$\sum_{i=1}^{N} \sum_{k=1}^{K} x_{ijk} = 1, j = 2,...N \quad (2)$$

$$\sum_{j=1}^{N} \sum_{k=1}^{K} x_{ijk} = 1, i = 2,...N \quad (3)$$

$$\sum_{i \in N} X_{i0k} = 1, \forall k \in K \quad (4)$$

$$\sum_{j \in N} X_{0jk} = 1, \forall k \in K \quad (5)$$

$$\sum_{i \in N} X_{iuk} - \sum_{j \in N} X_{ujk} = 0, \forall k \in K, \forall u \in N \quad (6)$$

$$X_{ijk} = 1 \Rightarrow y_{jk} = y_{ik} + q_i, \forall i, j \in N; \forall k \in K \quad (7)$$

$$Q_k \geq y_{ik} \geq 0, \forall i \in N; \forall k \in K \quad (9)$$

$$D_w \leq D_v, \forall i \in N; \forall w \in N_i^+; \forall v \in N_i^- \quad (10)$$

$$D_0 = 0 \quad (11)$$

$$X_{ijk} = 1 \Rightarrow e_i \leq A_i \leq l_i, \forall i, j \in N; \forall k \in K \quad (12)$$

$$X_{ijk} = 1 \Rightarrow e_i \leq A_i + s_i \leq l_i, \forall i, j \in N; \forall k \in K \quad (13)$$

$$X_{ijk} = 1 \Rightarrow D_i + t_{ijk} \leq (l_j - s_j), \forall i, j \in N; \forall k \in K \quad (14)$$

Les équations (2) et (3) assurent que chaque nœud n'est servi qu'une seule fois par un et un seul véhicule.
Les équations (4) et (5) assurent le non dépassement de la disponibilité d'un véhicule.
Un véhicule ne sort du dépôt et n'y revient qu'une seule fois.
L'équation (6) assure la continuité d'une tournée par un véhicule : le nœud visité doit impérativement être quitté.
Les équations (7), (8) et (9) assurent le non dépassement de la capacité de transport d'un véhicule.
Les équations (10) et (11) assurent le respect des précédences.
Les équations (12), (13) et (14) assurent le respect des fenêtres de temps. [9] [Kammarti, R et al]

### IV. Algorithme génétique pour la minimisation du coût total de transport

Au niveau de cette partie, nous allons présenter notre approche basée sur les algorithmes génétiques permettant de minimiser le coût total de transport.

#### A. Codage du chromosome

Le codage permet de représenter les solutions sous forme de chromosomes. Un chromosome est une suite (permutation) de nœuds, qui indique l'ordre dans lequel un véhicule doit visiter tous les nœuds. Ce type de codage (Figure 2) est appelé codage par liste de permutation.

| Nœud (i) | 0 | 5 | 2 | 6 | 4 | 3 | 10 | 8 | 7 | 9 | 1 | 0 |

**Figure 2 : Codage par liste de permutation**

Le nœud « 0 » représente le dépôt.

#### B. Décodage du passage des véhicules

Le décodage permet, à partir de chaque chromosome de la population, d'obtenir une solution initiale indiquant le passage de chaque véhicule sur les nœuds correspondants. (Figure 3)

| $V_1$ | $C_1$ | 0 | 2 | 4 | 1 | 5 | 0 |
|---|---|---|---|---|---|---|---|
| $V_2$ | $C_2$ | 0 | 3 | 6 | 7 | 10 | 0 |
| $V_3$ | $C_3$ | 0 | 8 | 9 | 0 | | |

**Figure 3 : Ordre de passage des véhicules**

## C. Croisement

Suite à la génération de la population initiale aléatoirement, nous procédons la phase de croisement qui assure la recombinaison de gênes parentaux pour former de nouveaux descendants. Pour ce faire, nous considérons $p$ la position de croisement, et nous traitons cette phase comme le montre la figure 4.

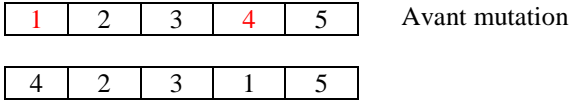

**Figure 4 : Croisement à un point**

## D. Mutation

La mutation (Figure 5) consiste à choisir deux allèles, au hasard, dans un chromosome et à échanger leurs valeurs respectives.

| 1 | 2 | 3 | 4 | 5 |   Avant mutation
|---|---|---|---|---|
| 4 | 2 | 3 | 1 | 5 |

**Figure 5 : Mutation de deux allèles quelconques**

Notons ici qu'il faut respecter les contraintes de précédences, afin de garantir qu'un client ne soit pas visité avant son fournisseur, et de capacité, pour assurer le non surcharge de chaque véhicule.

### E. Procédure de l'approche proposée pour la minimisation du coût total de transport

#### 1. Génération de la population initiale

Le mécanisme de génération de la population initiale permet de produire une population d'individus qui servira de base pour les futures générations. Le choix de la population initiale est important car il peut rendre plus ou moins rapide la convergence vers l'optimum global.

Dans notre cas, nous allons générer deux types de populations.

Une première population notée $P_{noeud}$, qui représente tous les nœuds à visiter par l'ensemble des véhicules, selon le codage par liste de permutation. (Figure 2)

La deuxième population notée $P_{vehicule}$ indique le nombre de nœuds visités par chaque véhicule. Sachant que $k$ varie entre 1 et $\frac{N'}{2}$ véhicules. La figure 6 montre un exemple d'un individu de $P_{vehicule}$, avec $N'$ =10.

| $V_1$ | $V_2$ | $V_3$ | $V_4$ | $V_5$ |
|---|---|---|---|---|
| 6 | 4 | 0 | 0 | 0 |

**Figure 6 : Exemple d'individu de $P_{vehicule}$**

#### 2. Génération de la population nœuds / Véhicules

Avant d'entamer la construction de la population $P_{noeud/vehicule}$, nous procédons à la phase de correction de précédence et de capacité entre les nœuds.

Soient les couples client / fournisseur suivants : (1,5), (2,8), (9,7), (10,3) et (4,6), notant que $Q_{k\,max}$ = 60, nous présentons, respectivement, dans les figures 7 et 8 le principe de correction de précédence et de capacité.

| 0 | 3 | 2 | 6 | 8 | 1 | 4 | 5 | 9 | 10 | 7 | 0 |
|---|---|---|---|---|---|---|---|---|---|---|---|

*Avant correction*

| 0 | 3 | 8 | 2 | 6 | 5 | 1 | 4 | 7 | 9 | 10 | 0 |
|---|---|---|---|---|---|---|---|---|---|---|---|

*Après correction*

**Figure 7 : principe de correction de précédence**

| 0 | 5 | 8 | 7 | 3 | 1 | 2 | 4 | 9 | 6 | 10 | 0 |
|---|---|---|---|---|---|---|---|---|---|---|---|

q=80

*Avant correction*

| 0 | 5 | 8 | 7 | 9 | 3 | 1 | 2 | 4 | 6 | 10 | 0 |
|---|---|---|---|---|---|---|---|---|---|---|---|

*Après correction*

**Figure 8 : principe de correction de capacité**

En considérant la population $P_{noeud}$, donnée par la figure 2, les procédures de correction et la population $P_{vehicule}$, donnée par la figure 6, nous illustrons dans la Figure 9 les individus correspondants de la population $P_{noeud/vehicule}$.

| $V_1$ | $C_1$ | 0 | 5 | 8 | 2 | 6 | 4 | 3 | 0 |
|---|---|---|---|---|---|---|---|---|---|
| $V_2$ | $C_2$ | 0 | 10 | 7 | 9 | 1 | 0 | | |

**Figure 9 : Exemple d'individus de $P_{noeud/vehicule}$**

#### 3. Procédure de calcul

Chaque véhicule part servir les nœuds dans sa tournée selon l'ordre d'affectation. Une fois rentré au dépôt, nous calculons le coût de transport correspondant à la distance parcourue, et nous répétons ceci jusqu'à ce que tous les nœuds soient servis.

Nous reproduisons ce travail pour chaque individu de la population $P_{vehicule}$ en tenant compte des différentes combinaisons possibles entre $P_{vehicule}$ et $P_{noeud}$, afin d'obtenir par la suite l'individu correspondant de la population $P_{noeud/vehicule}$ qui minimise notre fonction objectif.

La figure 10 représente la démarche à suivre pour la minimisation du coût total de transport.

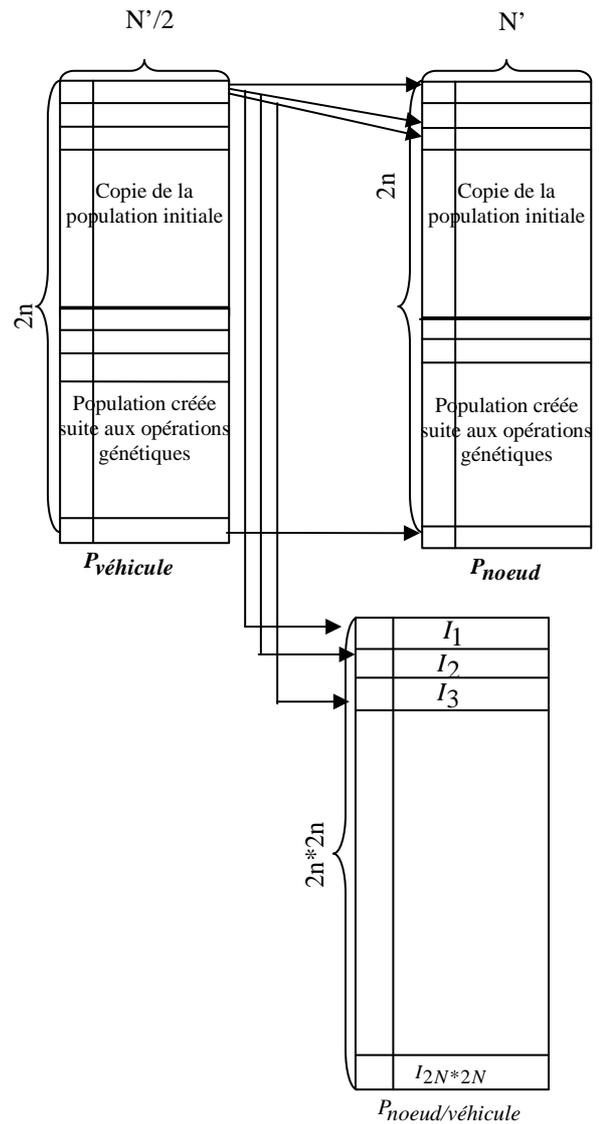

**Figure 10 : Procédure de calcul**

Avec :

n : taille de la population initiale,

$f_i$ : fonction objectif à minimiser (donnée par l'équation 1) correspondant à l'itération i ; i=1…2n*2n.

Nous déterminons par la suite :
$$f_{recherchée} = \min_{i} f_i \qquad (15)$$

### 4. Simulation

Nous représentons dans le tableau 1, les résultats de la simulation, donnés par notre approche.

| $N'$ | $n$ | $k$ | $Min\left(\sum_{k \in K}\sum_{i \in N}\sum_{j \in N} d_{ij} X_{ijk}\right)$ | $\min_{i} f_i$ |
|---|---|---|---|---|
| 20 | 100 | 2 | 878,84 | 59425,95 |
| 20 | 500 | 2 | 843,39 | 52296,84 |
| 50 | 100 | 2 | 2379,22 | 145534,61 |
| 50 | 500 | 2 | 2234,06 | 140743,17 |
| 100 | 100 | 4 | 4581,24 | 265197,12 |
| 100 | 500 | 4 | 4439,64 | 257383,70 |

**Tableau 1 : Résultats et Simulation**

## V. Conclusion

Dans cet article, nous avons proposé, suite à une étude des différentes approches proposées pour la résolution du 1-PDPTW et du m-PDPTW, un algorithme génétique pour la minimisation du coût total de transport. Pour cela, nous avons présenté la modélisation mathématique du problème étudié, et nous avons détaillé par la suite la démarche à suivre ainsi que, le processus de calcul pour déterminer la solution optimale, minimisant notre fonction objectif.